\documentclass[journal]{IEEEtran}
\usepackage{cite}
\usepackage{generic}
\usepackage{amsmath,amssymb,amsfonts}
\usepackage{graphicx}
\usepackage{algorithm,algorithmic}
\usepackage[table,xcdraw,dvipsnames,svgnames,x11names]{xcolor}
\usepackage{hyperref}
\usepackage{threeparttable}
\usepackage[utf8]{inputenc}
\usepackage{newunicodechar}
\usepackage{newtxmath}
\newunicodechar{̀}{\`{}}

\usepackage{booktabs}
\usepackage{multirow}
\usepackage{adjustbox}

\usepackage{textcomp}

\begin{document}
\title{AD-DAE: \textbf{A}lzheimer's \textbf{D}isease Progression Modeling with Unpaired Longitudinal MRI using \textbf{D}iffusion \textbf{A}uto-\textbf{E}ncoders}

\author{Ayantika Das, Arunima Sarkar, Keerthi Ram, Mohanasankar~ Sivaprakasam, and for the Alzheimer's Disease Neuroimaging Initiative* \thanks{*Data used in preparation of this article were obtained from the Alzheimer's Disease Neuroimaging Initiative(ADNI) database adni.loni.usc.edu). As such, the investigators within the ADNI contributed to the design and implementation of ADNI or provided data but did not participate in analysis or writing of this report.
%
Ayantika Das, Arunima Sarkar, and Mohanasankar Sivaprakasam  are with the Department of Electrical Engineering, Indian Institute of Technology Madras (IITM), Chennai
600036, India. Keerthi Ram and Mohanasankar Sivaprakasam are with Sudha Gopalakrishnan Brain Centre (SGBC), IITM, Chennai 600036, India. Corresponding author: Ayantika Das (email: dasayantika486@gmail.com).}
}

\maketitle

\begin{abstract}
Generative modeling frameworks have emerged as an effective approach to capture high-dimensional image distributions from large datasets without requiring domain-specific knowledge, a capability essential for disease progression modeling. Recent generative approaches have attempted to capture progression by mapping images to a latent space and guiding representations to generate follow-up images from previous time points. However, these methods impose constraints on distribution learning, resulting in latent spaces with limited controllability for generating follow-up images without paired subject-specific longitudinal guidance.

In order to enable \textit{controlled} movements in the latent representational space and generate progression images from a previous time-point image without subject-specific guidance, we introduce a conditionable \textit{Diffusion Auto-encoder} framework that forms a compact latent space capturing high-level semantics and providing means to control generation. 
Our approach leverages this latent space to condition and apply \textbf{controlled shifts} to the representations of previous time-point images by \textbf{isolating} progression and subject identity information for generating follow-up images.
The shifts are implicitly guided by correlating with progression attributes and constraining to Alzheimer's disease specific regions, \textbf{without} paired longitudinal \textbf{guidance}. We validate the generations through image quality metrics, volumetric progression analysis, and downstream tasks in Alzheimer's disease datasets from different sources. This demonstrates the effectiveness of our approach for Alzheimer's progression modeling and longitudinal image generation.
\end{abstract}


\section{Introduction}
\label{introduction}
Anatomical changes in the brain across a longitudinal time span, exceeding the rate of change in normal aging, are a crucial indicator for neurodegenerative diseases, like Alzheimer's. Structural MRI provides radiological features to observe and assess the nature of the disease progression \cite{mckhann2011diagnosis},\cite{jack2024revised}. In clinical practice, longitudinal assessment is performed by acquiring an MRI image of a current time point (follow-up) and comparing it against a previous time point \cite{barnes2009meta}. These comparisons are performed by segmenting specific brain regions known to be involved in Alzheimer's, followed by quantifying volumetric differences, and regional changes applying voxel-based morphometry \cite{ashburner2000voxel}. 

In contrast to progression assessment, \textbf{progression modeling} involves learning time- and disease-state dependent changes as observed in a training dataset of longitudinal MRI images of multiple subjects (individuals). The model is expected to capture the joint distribution of the image, the age at imaging, and the disease state of all subjects within the dataset. Further, \textbf{generative models for progression} use such distribution learning and offer the capability to sample the possible follow-up image given a previous time-point image, conditioned on the disease state and age. This offers a unique possibility of predicting future time point images.

Generative modeling approaches like Variational auto-encoder (\textbf{VAE}), have been applied to capture image distributions and generate follow-up images across longitudinal time spans. VAEs map images to a latent representational vector space that is regularized towards a desired prior distribution. 
While this approach drives the latent space towards encoding image details, 
VAEs may result in image generations that may not completely preserve the subject identity vis-a-vis the input image. Therefore, prior works use an external modeling-based conditioning that learns disease trajectories either from group-level trends or paired longitudinal guidance \cite{sauty2022progression, marti2023mc}. 
%

Generative Adversarial Networks (\textbf{GAN}) have also been widely utilized to capture MRI brain image distribution and generate plausible future follow-up images.  Although supervision from subject-specific paired longitudinal data prevails, GAN-based methods \textbf{without} explicit \textbf{guidance} have also been effective.
These techniques utilize previous time-point image for follow-up image generation by introducing controlled changes correlated with progression attributes (age/ disease) used for conditioning \cite{xia2021learning},\cite{wang2023spatial},\cite{pombo2023equitable}. 

We aim to implement this capability of generative modeling without explicit longitudinal supervision, similar to GAN-based approaches. However, unlike GANs, which directly learn input–output distribution mappings, we intend to utilize an auto-encoding formulation that embeds subject images as points in a latent representational space. This representation allows disease-related changes to be expressed as \textit{movements} in latent space, enabling control over image generation. To further enable progression with precise control, the latent space is required to be organized into dedicated subspaces that separately encode disease-related information and subject specific details. Our desired approach is a generative framework that \textbf{captures} image distributions across multiple \textit{time points} and disease states within a \textbf{controllable} latent space, allowing progression to be generated without \textbf{explicit} longitudinal paired guidance.

The recent denoising \textbf{diffusion}-based generative approaches have been employed to model Alzheimer's disease progression. Among these diffusion-based approaches, the latent diffusion methods were primarily utilized to learn follow-up representations by directly integrating progression-related attributes~\cite{puglisi2024enhancing}. While effective, these approaches rely on latents from separate encoder–decoder models to capture the image distribution, limiting control during the image generation. In order to better model the image distribution while enabling a controllable latent space, we propose to introduce a Diffusion Auto-encoder (DAE), an image-diffusion–based framework for longitudinal disease progression modeling.


The \textbf{Diffusion Auto-encoder} (DAE) model consists of an image encoder and a diffusion decoder, where the diffusion decoding component learns to generate images from pure Gaussian noise, guided by the latent representations produced by the encoder \cite{preechakul2022diffusion}. This design results in encoded representations that are explicit and compact in nature, thereby facilitating the separation of information within the latent space \cite{hudson2024soda}. Such a formulation has motivated the application of DAE models to image transformation tasks without explicit paired guidance.
%
Our contributions can be summarized as:
\begin{enumerate}

    \item We introduce \textbf{AD-DAE}, a diffusion autoencoder–based framework that models Alzheimer's disease progression by enabling controlled latent movements through conditioning a dedicated subspace of the latent representations, thereby separating progression-related factors from subject-identity components.

    \item We devise a mechanism to model progression without explicit guidance from subject-specific longitudinal pairs, by enforcing controlled changes that (i) correlate with progression attributes and (ii) are confined to Alzheimer's disease–relevant anatomical regions.

    \item We validate the proposed approach on 1016 subjects from the ADNI and OASIS datasets by (i) quantitatively assessing generation quality and generalization, (ii) evaluating volumetric measures of disease-specific regions, and (iii) analyzing latent space organization.

\end{enumerate}

\begin{figure*}[tbh]
  \centering
  \includegraphics[width=1
  \linewidth]{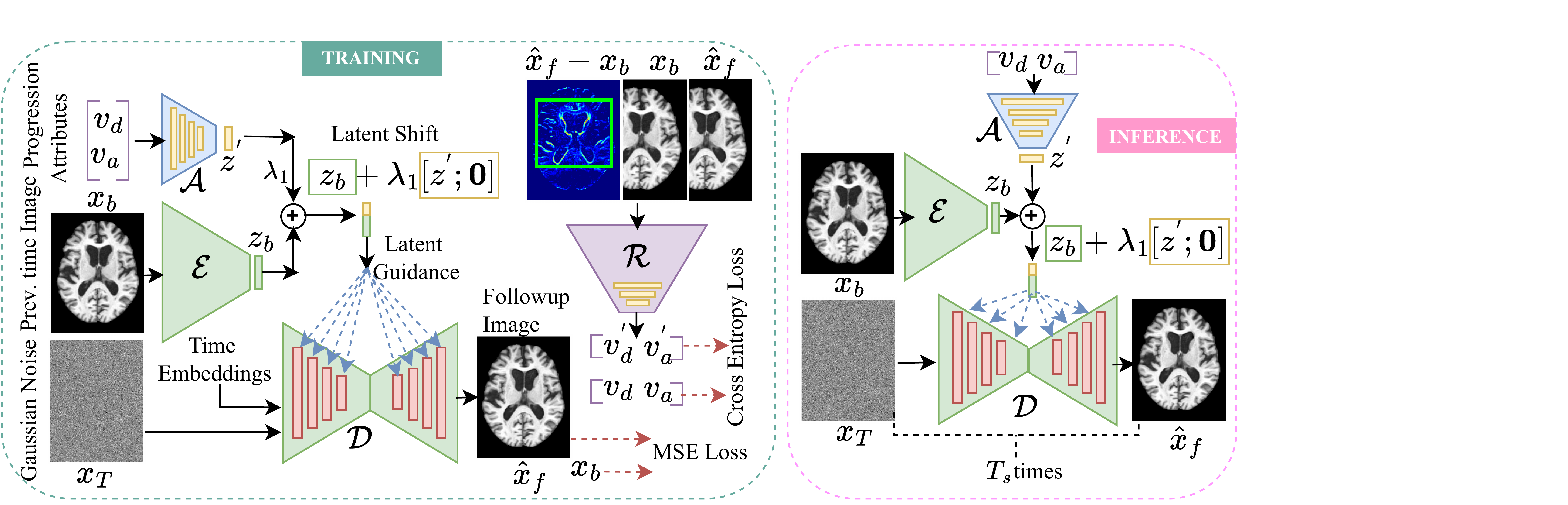}
\caption{From left to right, the training and inference strategy is described. \textbf{Training Module:} From left to right, the previous time-point image ($x_b$), Gaussian noise ($x_T$), time embeddings ($t$) and the progression attributes ($v_a,v_d$) are processed through the encoder ($\mathcal{E}$), latent shift module ($\mathcal{A}$) and the decoder ($\mathcal{D}$). The latent vector ($z_b$) from $\mathcal{E}$ gets shifted by $z^{'}$ incorporating progression factors and generating follow-up image ($\hat{x}_f$). The regression component $\mathcal{R}$ processes $x_b$, $\hat{x}_f$, and their residual to estimate the progression attributes, while optimizing $\mathcal{A}$. \textbf{Inference Module:} Image $x_b$, noise, and progression attributes are processed $T_s$ times to generate $\hat{x}_f$ with latent shift integration. }
\label{fig:architechture}
\end{figure*}
\section{Related Works}
Our approach can be categorized as a condition-driven diffusion-based progression modeling approach, trained without paired subject-specific guidance. Relating to this, we discuss: (i) Generative approaches for progression modeling, Diffusion methods (ii) for progression generation, and (iii) unpaired image-to-image translation.

\textit{\textbf{Generative Approaches for Longitudinal Data Generation}}: Generative frameworks conditioned on progression attributes and guided by paired longitudinal data have been widely adopted for progression modeling. GAN-based methods such as mi-GAN~\cite{zhao2020prediction}, 4D-DANI-Net~\cite{ravi2022degenerative}, and Identity-cGAN~\cite{jung2021conditional} incorporate biologically informed and identity-preserving constraints to model progression trajectories, while VAE-based approaches like DE-CVAE~\cite{he2024individualized} employ dual-encoder designs to capture progression attributes under paired longitudinal guidance.
Among unpaired approaches, CAAE~\cite{zhang2017age} demonstrated that latent space traversals can generate age-progressed images. IPGAN~\cite{xia2021learning} and Identity-3D-cGAN~\cite{jung2023conditional} incorporate identity-preserving constraints while correlating generation with progression attributes. Other methods, like CounterSynth~\cite{pombo2023equitable} and SITGAN~\cite{wang2023spatial}, apply diffeomorphic transformations guided by progression attributes. While these unpaired approaches are effective, in order to precisely capture temporal information, better modeling capabilities are required.

\textit{\textbf{Diffusion Models for Longitudinal Data Modeling}}: Denoising diffusion models have been applied to progression modeling in both latent and image spaces. \textit{Latent diffusion} models like BrLP~\cite{puglisi2024enhancing} generate follow-up latent representations by conditioning with progression attributes and guidance from paired follow-ups. Similarly, MRExtrap~\cite{kapoor2024mrextrap} assumes linear age-dependent trajectories to generate follow-up latents.
Although latent diffusion provides a compact conditioning space, its performance is limited by the representational capacity of a separate encoder–decoder. In contrast, \textit{image-level diffusion} directly models in image space, enabling more precise image distribution modeling. Methods such as SADM~\cite{yoon2023sadm} encode temporal dependencies using transformer-based conditioning, while TADM~\cite{litrico2024tadm} predicts additive residuals required to generate follow-ups, through paired guidance.

\textit{\textbf{Diffusion Models for Unpaired Image-to-Image Tasks}}: Latents from diffusion models have been leveraged for multiple image translation tasks without paired guidance. Diffusion Deformable Model~\cite{kim2022diffusion} and DiffuseMorph~\cite{kim2022diffusemorph} generate target images by latent transformations using reference target without the usage of original target \cite{dalva2024noiseclr}. However, these latents lie in time-dependent image space, which limits controllability \cite{dhariwal2021diffusion}. To address this, Diffusion Auto-encoders (DAE)~\cite{preechakul2022diffusion, zhang2022unsupervised} introduce a compact, structured latent space~\cite{das2025posdiffae} directly guiding the denoising diffusion process~\cite{ho2020denoising}, \cite{nichol2021improved}.
DAE latent representations have been applied to diverse image translation tasks \cite{huang2025diffusion}. Methods such as SSR-Encoder \cite{zhang2024ssr} and PADA \cite{li2023pluralistic} integrate textual guidance into DAE latents, and HDAE \cite{lu2024hierarchical} combines latents from multiple references to generate target images without explicit guidance from ground truth targets. Further studies show that DAE latents can isolate attributes across latent subsets \cite{yueexploring}, \cite{hudson2024soda}. Building on these insights, we utilize the DAE latent space for modeling disease progression.

\section{Methodology}
\label{sec:methods}

Our proposed approach AD-DAE is a Diffusion Auto-encoder (DAE)-based framework which generates disease progression images through: (i) \textit{Latent Shift} module, which induces controlled movements in the latent representational space of DAE to generate progression follow-up images from previous time-point; and (ii) \textit{Consistency} module, which implicitly guides the generation of follow-up images by correlating with progression attributes and constraining the generations to be confined to Alzheimer's disease–relevant anatomical regions. 
%
The architectural flow of our devised method is illustrated in Figure~\ref{fig:architechture}.

In the \textbf{DAE} formulation, the encoding ($\mathcal{E}$) component maps input images ($x \in \mathbb{R}^{h \times w}$) into latent representations ($z \in \mathbb{R}^d$), which guides ($z = \mathcal{E}(x)$) the denoising diffusion component ($\mathcal{D}$) during image generation ($\hat{x} \in \mathbb{R}^{h \times w}$). This latent space enables us to devise a mechanism to generate disease progression images by shifting the progression-related dimensions of the DAE latent representation ($z$).

The latent movement required to generate a progression follow-up image ($x_f$) from a subject's previous time-point image ($x_b$) is estimated through a \textbf{Latent Shift} module ($\mathcal{A}$) that maps progression attributes (cognitive status and age) into a latent shift vector ($z^{'}$). This mapping ensures that the latent shift $z^{'}$ lies in the subspace of the latent representational space ($z = \mathcal{E}(x)$) learnt by DAE. The latent shift operation is constrained to allow decoupling of progression from subject-identity related factors of latent $z_b$. The resulting shifted latent representation further guides $\mathcal{D}$ to generate a follow-up image ($\hat{x}_{f}$). 

The \textbf{Consistency} module ($\mathcal{R}$) regresses progression attributes (inputs to $\mathcal{A}$) from the previous time-point $x_b$ and the generated follow-up $\hat{x}_f$ images, by focusing on changes in the disease-specific regions of the images. This ensures that changes introduced within $\hat{x}_f$ are guided in terms of (i) direction and magnitude through the progression attributes and (ii) spatial locations through disease-related anatomy masks.  
This module ($\mathcal{R}$) implicitly guides $\mathcal{A}$ to produce meaningful latent movements and DAE to generate $\hat{x}_{f}$ by transforming $x_b$ in disease-specific regions.

The following subsections detail the components of \textbf{AD-DAE}, (i) Diffusion Auto-encoder (Subsection~\ref{subsec:DAE}), (ii) Latent Shift Estimation Module (Subsection~\ref{subsec:PEM}), (iii) Consistency Module (Subsection~\ref{subsec:RM}), along with the (iv) Training and Inference process (Subsection~\ref{subsec:Train_infer}).

\subsection{Diffusion Autoencoder (DAE)}
\label{subsec:DAE}

At the core of our progression modeling approach is a diffusion-based auto-encoder, which has an encoding component ($\mathcal{E}$) and a denoising diffusion decoding component ($\mathcal{D}$). 

\subsubsection{Denoising Decoder ($\mathcal{D}$)}
The denoising decoder $\mathcal{D}$ is an image-based diffusion that iteratively learns to transform a noise sample (\texorpdfstring{$x_T$}{xT}), drawn from a standard Gaussian distribution, into a target image (\texorpdfstring{$x_0$}{x0}) utilizing the DDIM \cite{song2020denoising} strategy.
This iterative transformation is the reverse of the forward diffusion process given by a time-dependent Gaussian distribution as, 
\begin{equation}
    \label{eq:q_xt_xt-1}
   q(x_t|x_{t-1}) = \mathcal{N}(\sqrt{1 - \beta_t} x_{t-1}, \beta_t I) 
\end{equation}
where $\beta_t$ represents the noise levels. The diffusion process spans across a total of $T$ time steps. At each time step $t$, the goal of $\mathcal{D}(x_{t},t)$ is to model the reverse process ($p_{\phi}(x_{t-1}|x_{t})$) given the noisy input $x_{t}$. This noisy image can be approximated using the following distribution,
\begin{equation}
    \label{eq:q_xt_x0}
   q(x_t|x_0) = \mathcal{N}(\sqrt{\alpha_t} x_0, (1 - \alpha_t)I)  
\end{equation} 
where $ \alpha_t = \prod_{s=1}^{t} (1 - \beta_s) $ and $x_t = \sqrt{\alpha_t } x_0 + \sqrt{1 - \alpha_t} \epsilon$, $\epsilon \sim \mathcal{N}(0, I)$. The model $\mathcal{D}$ is configured to directly estimate $x_0$ such that, $ \hat{x}_{0} = \mathcal{D}( {x}_t, t)$. 
With this configuration, the reverse process to be modeled is given by the distribution as,
\small
\begin{equation}
    \label{eq:p_xt-1_xt}
    \small
p_\phi\left({x}_{t-1} \mid  {x}_t \right)= \begin{cases}\mathcal{N}\left( \mathcal{D}\left( {x}_1, 1\right),  {0}\right) & \text { if } t=1 \\ q\left( {x}_{t-1} \mid  {x}_t,  \hat{x}_{0} \right) & \text { otherwise }\end{cases}
\end{equation}

where $q(x_{t-1} | x_t, \hat{x}_0)$ is derived from equation \ref{eq:q_xt_xt-1} and \ref{eq:q_xt_x0}, and is given as,
\begin{equation}
    \label{eq:q_xt-1_xt_x0}
    q(x_{t-1} | x_t, \hat{x}_0) = \mathcal{N} (\sqrt{\alpha_{t-1}} \hat{x}_0  + \sqrt{1 - \alpha_{t-1}} (\frac{x_t - \sqrt{\alpha_t} \hat{x}_0 }  {\sqrt{1 - \alpha_t}}) ,0 ) 
    \end{equation} 

This formulation allows the reverse denoising process to sample $x_{t-1}$ given $x_{t}$ in a tractable manner. The entire pipeline is implemented through a convolution UNet \cite{ho2020denoising}, \cite{song2020denoising}. This UNet, in the denoising process, serves as a decoder in the DAE module, being conditioned on latent representations from the integrated encoder ($\mathcal{E}$), which is detailed below.


\subsubsection{Guiding Encoder ($\mathcal{E}$)} 
\label{subsubsec:encoder}
The denoising decoder (\texorpdfstring{$\mathcal{D}$}{D}) is guided by the latent representation produced by the encoder (\texorpdfstring{$z = \mathcal{E}(x_0)$}{z=E(x0)}), enabling an auto-encoding-like \cite{preechakul2022diffusion} structure (\texorpdfstring{$\hat{x}_0 =~ \mathcal{D}(x_t,t,z)$}{x0hat = D(xt,t,z)}) during the iterative image generation process. 
%
The guidance from the encoder ($\mathcal{E}$) is integrated into $\mathcal{D}$ by conditioning each of its layers through activation modulation. These activations within $\mathcal{D}$ are first normalized via group normalization and then modulated using learned scale and bias parameters derived from $z$.

The latent space offered by the encoder $\mathcal{E}$ enables image transformation without paired supervision, since the latent representations ($z$) can encapsulate meaningful information. 
This can be attributed to the denoising process that specializes in reconstructing \textbf{high-frequency} details, allowing the encoded latent representation to focus on capturing high-level \textbf{semantic} structures \cite{hudson2024soda}. In contrast, conventional auto-encoders require the latent space to encode both low-level details and high-level semantics.

By decoupling these factors, our framework enables the latent representations to emphasize within structural regions, such as the ventricles and hippocampus, which are affected by Alzheimer's progression. In order to introduce changes in these regions and generate progression imaga es, shift is incorporated in the latent representation of DAE through the latent shift estimation module $\mathcal{A}$ as detailed below.



\subsection{Latent Shift Estimation Module}
\label{subsec:PEM}
The latent shift estimation module $\mathcal{A}$ estimates the shift required to generate a progression follow-up image ($\hat{x}_{f}$) of a previous time-point image $x_b$. The latent shift ($z^{'}$) is introduced through an additive operation on the latent representation $z_b$ ($z_b=\mathcal{E}(x_b)$) to estimate the follow-up latent representation ($z^{'}_{f}$). This estimated representation is employed to predict the follow-up image given by $\hat{x}_f =~ \mathcal{D}(x_t,t,z^{'}_{f})$.

\textit{Input Attributes:} The shift estimation module $\mathcal{A}$ takes the progression attributes, (i) age gap (${v}_a$) between the previous time-point and follow-up, and (ii) the cognitive status (${v}_d$) as input. The progression attributes are represented using, $(v_{d}^{k_{i}}, v_{a}^{k_{j}}) = ((0, .., 1_{k_{i}}, .., 0),(0, \dots, 1_{k_{j}}, .., 0))$, where $k$ in $v^{k_{i}}_d$ and $v^{k_{j}}_a$ indicates the positions indicating each of the cognitive sub-types ($k_{i}$) and age-gaps ($k_{j}$) respectively. The three cognitive statuses are CN, MCI, and AD. The age gaps and maximum allowable age gaps can be chosen as compatible with the training set.


The latent shift module processes these progression attributes and produces the required latent shift $z^{'} = \mathcal{A}(v_d, v_a)$, which captures cognitive and age-related movements in the latent space of DAE. The shift vector ($z^{'} \in \mathbb{R}^m, m<d$) is added to a subset of the latent $z_b$ through the following operation, $z_{f}^{'} = z_{b} + [z^{'}; \mathbf{0}]$, where $z^{'} \in \mathbb{R}^m$ and $\mathbf{0} \in \mathbb{R}^{d - m}$.
This shift in the first $m$ elements of the $d$ latent dimensions of $z_{b}$ ensures progression-relevant attributes are \textbf{decoupled} from the subject-specific identity features encoded in the remaining $d-m$ dimensions. This isolation of progression-specific properties enables the generated follow-up image ($\hat{x}_f$) to reflect structural changes, such as enlargement or shrinkage, associated with progression, while preserving identity-relevant anatomical structures.

\begin{table}[!b]
\caption{Parameter Details}
\begin{threeparttable}
\centering
\resizebox{\columnwidth}{!}{%
\begin{tabular}{@{}cc|cc@{}}
\toprule
\textbf{Parameters} &
  \textbf{Values} &
  \textbf{Parameters} &
  \textbf{Values} \\ \midrule
Epochs &
  \begin{tabular}[c]{@{}c@{}}Auto-encode: 50\\ Progression: 100\end{tabular} &
  \begin{tabular}[c]{@{}c@{}}Latent \\ Dimensions \\ ($d, m$)\end{tabular} &
  512, 50 \\ \midrule
\begin{tabular}[c]{@{}c@{}}Axial Height ($H$), \\ Width ($W$), \\ Slices ($D$)\end{tabular} &
  \begin{tabular}[c]{@{}c@{}}208,\\ 160,\\ 100\end{tabular} &
  $\lambda_1$ &
  \begin{tabular}[c]{@{}c@{}}Auto-encode: 0 \\ Progression: 1\end{tabular} \\ \midrule
Optimizer &
  Adam &
  $\lambda_2$, $\lambda_3$ &
  1, 0.001 \\ \midrule
Learning Rate &
  0.001 &
  \begin{tabular}[c]{@{}c@{}}Cognitive Types: $v_{d}^{k_{i}}$ \\ $(0, .., 1_{k_{i}}, .., 0)$\end{tabular} &
  $k_{i} \in \{1,2,3\}$ \\ \midrule
\begin{tabular}[c]{@{}c@{}}Diffusion Noise \\ ($\beta_t$, $T$, $T_s$)\end{tabular} &
  \begin{tabular}[c]{@{}c@{}}Linear \\ Scheduling,\\ 1000, 50\end{tabular} &
  \begin{tabular}[c]{@{}c@{}}Age Gap: $v_{a}^{k_{j}}$\\ $(0, .., 1_{k_{j}}, ., 0)$\end{tabular} &
  $k_{j} \in \{4,..,13\}$ \\ \midrule
$\mathcal{E}$, $\mathcal{D}$ &
  \begin{tabular}[c]{@{}c@{}}ResNet, \\ UNet\cite{preechakul2022diffusion}\end{tabular} &
  \begin{tabular}[c]{@{}c@{}}Region \\ Segmentation ($r$)\end{tabular} &
  SynthSeg\cite{billot2023synthseg} \\ \midrule
$\mathcal{R}$ &
  \begin{tabular}[c]{@{}c@{}}ResNet,\\ MLP layers\end{tabular} &
  \begin{tabular}[c]{@{}c@{}}Jacobian-\\ based Analysis\end{tabular} &
   ANTs\tnote{1}\\ \midrule
$\mathcal{A}$ &
  \begin{tabular}[c]{@{}c@{}}MLP Layers,\\ Activations\end{tabular} &
  \begin{tabular}[c]{@{}c@{}}Disease \\ Classification\end{tabular} &
  \begin{tabular}[c]{@{}c@{}}ResNeXt\cite{xie2017aggregated},\\ MLP Layers\end{tabular} \\ \bottomrule
\end{tabular}%
}
\begin{tablenotes}
\item[1] \href{https://antspy.readthedocs.io/en/latest/_modules/ants/registration/create_jacobian_determinant_image.html}{ANTs}
\end{tablenotes}
\end{threeparttable}
\label{table:params}
\end{table}

\begin{table*}[!t]
\caption{Quantitative evaluation of AD-DAE against baseline methods on the \textit{Test Set}, using PSNR, SSIM, and MSE. Model size and inference time per 3D image ($X$) are also reported. $^*$ indicates statistical significance ($p<0.01$).}
\centering
\resizebox{2\columnwidth}{!}{%
\begin{tabular}{@{}ccccccccc@{}}
\toprule
\textbf{Methods} &
  \multicolumn{2}{c}{\textbf{PSNR ($\uparrow$)}} &
  \multicolumn{2}{c}{\textbf{SSIM ($\uparrow$)}} &
  \multicolumn{2}{c}{\textbf{MSE ($\downarrow$)}} &
  \textbf{\begin{tabular}[c]{@{}c@{}}Model \\ Size\end{tabular}} &
  \textbf{\begin{tabular}[c]{@{}c@{}}Inference \\ Time per X\end{tabular}} \\ \midrule
\textbf{}                          & \textbf{CN}  & \textbf{MCI/ AD} & \textbf{CN}  & \textbf{MCI/ AD} & \textbf{CN}    & \textbf{MCI/ AD} & \textbf{(MB)} & \textbf{(s)} \\ \midrule
Naive Baseline                     & 27.25 $\pm$ 2.12 & 26.75 $\pm$ 2.07     & 0.93 $\pm$ 0.021 & 0.92 $\pm$ 0.021     & 0.0021 $\pm$ 0.001 & 0.0024 $\pm$ 0.001   & -             & -            \\ \midrule
CAAE\cite{zhang2017age}            & 21.21 $\pm$ 0.73 & 21.03 $\pm$ 0.80     & 0.53 $\pm$ 0.033 & 0.52 $\pm$ 0.035     & 0.0077 $\pm$ 0.001 & 0.0080 $\pm$ 0.001   & 52.30         & 2.56        \\ \midrule
IPGAN\cite{xia2021learning}        & 25.86 $\pm$ 2.12 & 25.31 $\pm$ 2.13     & 0.92 $\pm$ 0.032 & 0.91 $\pm$ 0.034     & 0.0030 $\pm$ 0.001 & 0.0034 $\pm$ 0.002   & 52.30         & 2.34         \\ \midrule
UVCGAN\cite{torbunov2023uvcgan}    & 26.43 $\pm$ 2.57 & 25.13 $\pm$ 1.53     & 0.92 $\pm$ 0.033 & 0.91 $\pm$ 0.031     & 0.0025 $\pm$ 0.001 & 0.0041 $\pm$ 0.001   & 122.90        & 5.24         \\ \midrule
BrLP\cite{puglisi2024enhancing}    & 26.71 $\pm$ 1.02 & 26.20 $\pm$ 1.14     & 0.79 $\pm$ 0.022 & 0.79 $\pm$ 0.025     & 0.0029 $\pm$ 0.001 & 0.0030 $\pm$ 0.001   & 576.76        & 13.38        \\ \midrule
DE-CVAE\cite{he2024individualized} & 27.32 $\pm$ 2.98 & 26.99 $\pm$ 2.83     & 0.65 $\pm$ 0.090 & 0.63 $\pm$ 0.082     & 0.0023 $\pm$ 0.001 & 0.0024 $\pm$ 0.001   & 648.12        & 3.89         \\ \midrule
SITGAN\cite{wang2023spatial}       & 28.73 $\pm$ 3.25 & 28.09 $\pm$ 3.23     & \textbf{0.94 $\pm$ 0.033} & 0.93 $\pm$ 0.034     & 0.0019 $\pm$ 0.001 & 0.0022 $\pm$ 0.001   & 98.78         & 5.08         \\ \midrule
AD-DAE                             & \textbf{30.10$^*$ $\pm$ 3.05} & \textbf{29.43$^*$ $\pm$ 3.14}     & \textbf{0.94$^*$ $\pm$ 0.033} & \textbf{0.94$^*$ $\pm$ 0.031}     & \textbf{0.0017$^*$ $\pm$ 0.001} & \textbf{0.0019$^*$ $\pm$ 0.001}   & 129.18        & 10.03        \\ \bottomrule
\end{tabular}%
}
\label{table:quanti}
\end{table*}

\subsection{Consistency Module}
\label{subsec:RM} 
The inputs to the consistency module ($\mathcal{R}$) are: (i) previous time-point image ($x_b$) (ii) predicted follow-up image ($\hat{x}_{f}$) and (iii) their residual ($x_b - \hat{x}_{f}$), to regress the progression attributes, $(v_d^{'}, v_a^{'}) = \mathcal{R}(x_{b}, \hat{x}_{f}, x_b - \hat{x}_{f})$. 
The module $\mathcal{R}$ focuses on the changes introduced between $x_{b}$ and $\hat{x}_{f}$ and regresses the progression attributes $(v_d^{'}, v_a^{'})$. This encourages transformation of $x_b$ to $\hat{x}_f$ and enables the \textit{direction and magnitude} of the changes to be \textbf{correlated} to progression attributes. 


The changes generated during the transformation of $x_b$ to $\hat{x}_f$ need to be \textit{spatially constrained} to ensure that \textbf{disease-specific} changes are introduced while preserving subject identity. This is achieved by masking the residual $x_b - \hat{x}_f$ to retain changes only within certain progression-relevant regions like ventricles, hippocampus, and amygdala. These regions are selected based on their disease-specific relevance, since in MCI, and early AD, some of the earliest and most reproducible MRI-observable changes occur in medial temporal lobe structures such as the hippocampus and amygdala, while lateral ventricular enlargement reflects downstream tissue loss and global atrophy progression \cite{planche2022structural},\cite{da2017neuro}.

\textit{Residual Mask}: The residual mask is generated by segmenting progression-related regions ($r$) from the previous time-point image $x_b$ using the 3D SynthSeg \cite{billot2023synthseg}. The masks are morphologically dilated with a window size of $5$ to account for potential regional expansion in $\hat{x}_f$. An aggregated bounding box is extracted from the masks that are applied to the residual between $x_b$ and $\hat{x}_f$, ensuring sensitivity to segmentation is reduced.

By enforcing the notion of direction and magnitude under spatial constraints during regression, the model facilitates meaningful generation of $\hat{x}_f$, thereby ensuring that the output $z^{'}$ from $\mathcal{A}$ induces a controlled shift of $z_b$ in the required direction and with appropriate magnitude, ($z_b + [z^{'};\mathbf{0}]$). The regression loss governing this optimization is detailed in Subsection~\ref{subsec:Train_infer}.

\subsection{Training And Inference}
\label{subsec:Train_infer} 
\subsubsection{Training Objectives Functions} The DAE module of our model is primarily optimized with MSE loss function $\mathcal{L}_{{MSE}} = \| x_b - \hat{x}_{0} \|_{2}^{2}$, where $\hat{x}_{0} = \mathcal{D}(x_t,t,z_{b} + \lambda_{1}[z^{'}; \mathbf{0}])$. The consistency module ($\mathcal{R}$) is optimized by a cross-entropy-based objective function given as, $\mathcal{L}_{\mathrm{CE}}~ =~ - \mathbf{v}_d^\top \log \mathbf{v}_d'~ -~ \mathbf{v}_a^\top \log \mathbf{v}_a'$. The overall objective function is given as,
\begin{equation}
    \mathcal{L} = 
    \begin{cases}
    \mathcal{L}_{MSE} , & \text{if } \lambda_1 = 0 \\
    \lambda_2 \mathcal{L}_{MSE} + \lambda_3 \mathcal{L}_{CE} , & \text{if } \lambda_1 = 1
    \end{cases}
\end{equation}
The parameter $\lambda_{1}$ enables the model to alternate between two modes: auto-encoding ($\hat{x}_{0} = \hat{x}_b$) when $\lambda_1=0$, and progression generation ($\hat{x}_{0} = \hat{x}_f $) when $\lambda_1=1$. The model is trained in $\lambda_1=0$ mode for the initial few epochs to learn meaningful image features without conditions.

In progression generation mode, the loss term $\mathcal{L}_{CE}$ ensures that $\mathcal{R}$ regress attributes $(v_d^{'}, v_a^{'})$, while extracting information from disease-specific regions of $x_b - \hat{x}_f$ and implicitly guiding $\mathcal{A}$ to generate required latent shift $z^{'}$. Thus, generation of meaningful changes (required direction and magnitude) in specific spatial locations of $\hat{x}_{f}$ will lead to convergence of $\mathcal{R}$ and reduce sparsity in $z^{'}$.
While this supports unpaired training, $\mathcal{L}_{MSE}$ facilitates both meaningful reconstruction and identity preservation. The parameters $\lambda_2$ and $\lambda_3$ balance allowable transformations against these constraints.

\subsubsection{Inference}
During inference, the previous time-point representation $z_b = \mathcal{E}(x_b)$ is shifted by $z^{'}$ estimated from progression attributes $(\mathbf{v}_d,\mathbf{v}_a)$ to form $z_f^{'}$. This modified latent guides the denoising decoder $\mathcal{D}$ to generate the follow-up image $\hat{x}_f$. The decoder $\mathcal{D}(x_t,t,z_f^{'})$ iteratively samples $\hat{x}_f$ from $x_T$, using $z_f^{'} = z_b + [z^{'};\mathbf{0}]$ and following Eqs.~\ref{eq:p_xt-1_xt} and~\ref{eq:q_xt-1_xt_x0} for $T_s$ steps.

\section{Experimental Setup}


\subsection{Datasets}
We have utilized longitudinal brain MRI images from the Alzheimer's Disease Neuroimaging Initiative (ADNI) \href{https://adni.loni.usc.edu/}{(adni.loni.usc.edu)} \cite{jack2008alzheimer} and Open Access Series of Imaging Studies (OASIS) \href{https://www.oasis-brains.org/}{(oasis-brains.org)} \cite{lamontagne2019oasis} repositories. The images considered from both repositories were T1-weighted 3D images.
\textit{\textbf{ADNI:}} From the ADNI repository, we have considered subjects from three different cognitive statuses: cognitively normal (CN), mildly cognitively impaired (MCI), and AD. The images utilized were within the age range of 63 $-$ 87 years, with an average of $2.93 \pm 1.31$ images per subject acquired over an average longitudinal span of $2.89 \pm 0.79$ years (average image acquisition frequency is $11$ months per subject).
\textit{\textbf{OASIS:}} Similar to the ADNI from the OASIS repository, we have utilized data from the cognitively normal, MCI due to AD, and AD dementia groups. The age range of the images was within 60 $-$ 90 years with an average of $2.73 \pm 1.15$ images per subject acquired over an average longitudinal span of $3.13 \pm 1.52$ years (average image acquisition frequency is $14$ months per subject).


\begin{figure*}[]
  \centering
  \includegraphics[width=1
  \linewidth]{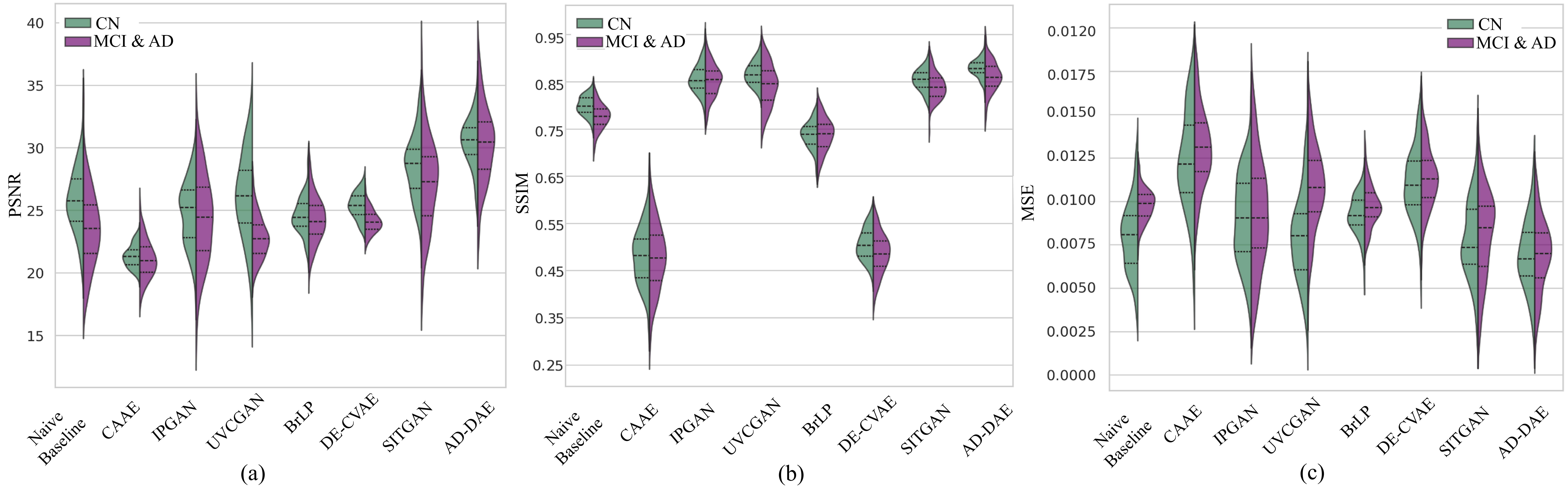}
\caption{Left to right: Performance comparison of AD-DAE on \textit{Cross-Data Setup} with baseline methods in terms of (a) PSNR, (b) SSIM, and (c) MSE. CN is shown in green and MCI \& AD in violet, represented with their mean and first-quartile values. Across all metrics, AD-DAE demonstrates relatively better performance.}
\label{fig:Oasis_plot}
\end{figure*}

\subsection{Evaluation Metrics}
\label{subsec:Evaluation Metrics}

\textbf{\textit{Image Quality Metrics:}} To evaluate the similarity between predicted follow-up ($\hat{x}_{f}$) and ground-truth (real data) follow-up ($x_{f}$), standard image-based measures, Peak Signal-to-Noise Ratio (PSNR), Structural Similarity Index (SSIM), and Mean Squared Error (MSE) are calculated. \textbf{\textit{Volumetric Analysis:}} To assess how well progression is modeled in 3D volumes, we compare errors between the volumetric estimates of generated and ground truth images. We measure region-wise Mean Absolute Error (MAE) between $V^{r}_{rel\_pred}=(V_{\hat{X}_f}^{r} - V_{{X}_b}^{r})/(V_{X_b}^{r})$ and $V^{r}_{rel\_gt}=(V_{X_f}^{r} - V_{{X}_b}^{r})/(V_{X_b}^{r})$, where $V_{X}^{r}$ is the sum of voxels belonging to region $r$ ($r \in $ Hippocampus/ Amygdala/ Lateral Ventricular) in any 3D image volume ($X$). The region-wise estimates are extracted by segmenting the 3D image volumes utilizing the approach in Table \ref{table:params}. 
We further measure the difference between the deformations required to transform the generated follow-up ($\hat{X}_{f}\rightarrow X_{b}$) and the ground-truth (real data) follow-up (${X}_{f}\rightarrow X_{b}$) to the previous time-point image utilizing Jacobian determinant (Table \ref{table:params}). 
\textit{\textbf{Clinical Evaluation:}} To clinically validate the generations, we have added qualitative interpretation of an expert from the neuroscience domain with 10$+$ years of experience.

\subsection{Implementation Details}

\subsubsection{Dataset Details}
From the ADNI dataset, we constructed a \textit{Train Set} of 486 subjects (179 CN, 160 MCI, 147 AD) and a \textit{Test Set} of 466 subjects (159 CN, 156 MCI, 151 AD). A \textit{Latent Swap Set} was created by pairing CN and AD subjects of similar age (within 0.5 years). In a pair, the age gap associated with the AD subject is of two years or more, and with the CN subject is less than two years. For generalization assessment, a \textit{Cross-Data Setup} was created with the unseen OASIS dataset, consisting of 550 subjects (330 CN, 137 MCI, 83 AD). 
A \textit{Cross-Cognition Set} was constructed from the ADNI dataset, with MCI subjects labeled as progressive, indicating conversion to AD. The last available MCI time-point images were chosen as input for AD generation.
All brain MRIs were pre-processed using skull stripping \cite{isensee2019automated}, affine registration to MNI space, and intensity normalization~\cite{shinohara2014statistical}.


\subsubsection{Model Details}
The models were implemented in PyTorch version 2.0.1 on an 80 GB NVIDIA A100 GPU. All the parameter details related to AD-DAE, and all downstream analyses are listed in Table \ref{table:params}. Full training cycle takes 9.5 hrs, with parameters presented in Table \ref{table:params} and batch size 10. The size of all the models and the inference time required to process an image are available in Table \ref{table:quanti}. The code-base is made available at \href{https://github.com/ayantikadas/AD_DAE}{https:/github.com/ayantikadas/AD\_DAE}.

\textit{Baseline Methods:} We compare against (i) \textbf{GAN}-based methods trained without paired guidance, including CAAE~\cite{zhang2017age} and IPGAN~\cite{xia2021learning}, which enforce subject-identity preservation, UVCGAN~\cite{torbunov2023uvcgan}, which adopts a transformer-based cycle-GAN formulation, and SITGAN~\cite{wang2023spatial}, which correlates image generation with age and cognitive status; (ii) latent \textbf{diffusion}-based BrLP~\cite{puglisi2024enhancing}, which integrates multiple progression attributes and uses subject-specific paired guidance; and (iii) \textbf{VAE}-based DE-CVAE~\cite{he2024individualized}, which employs a separate encoder for progression attributes and paired guidance. All baselines were implemented using official source codes available at \href{https://github.com/ayantikadas/AD_DAE}{https://github.com/ayantikadas/AD\_DAE}
 and evaluated with identical progression attributes (cognitive state, age, and region information).

%

\begin{figure*}[t]
  \centering
  \includegraphics[width=1
  \linewidth]{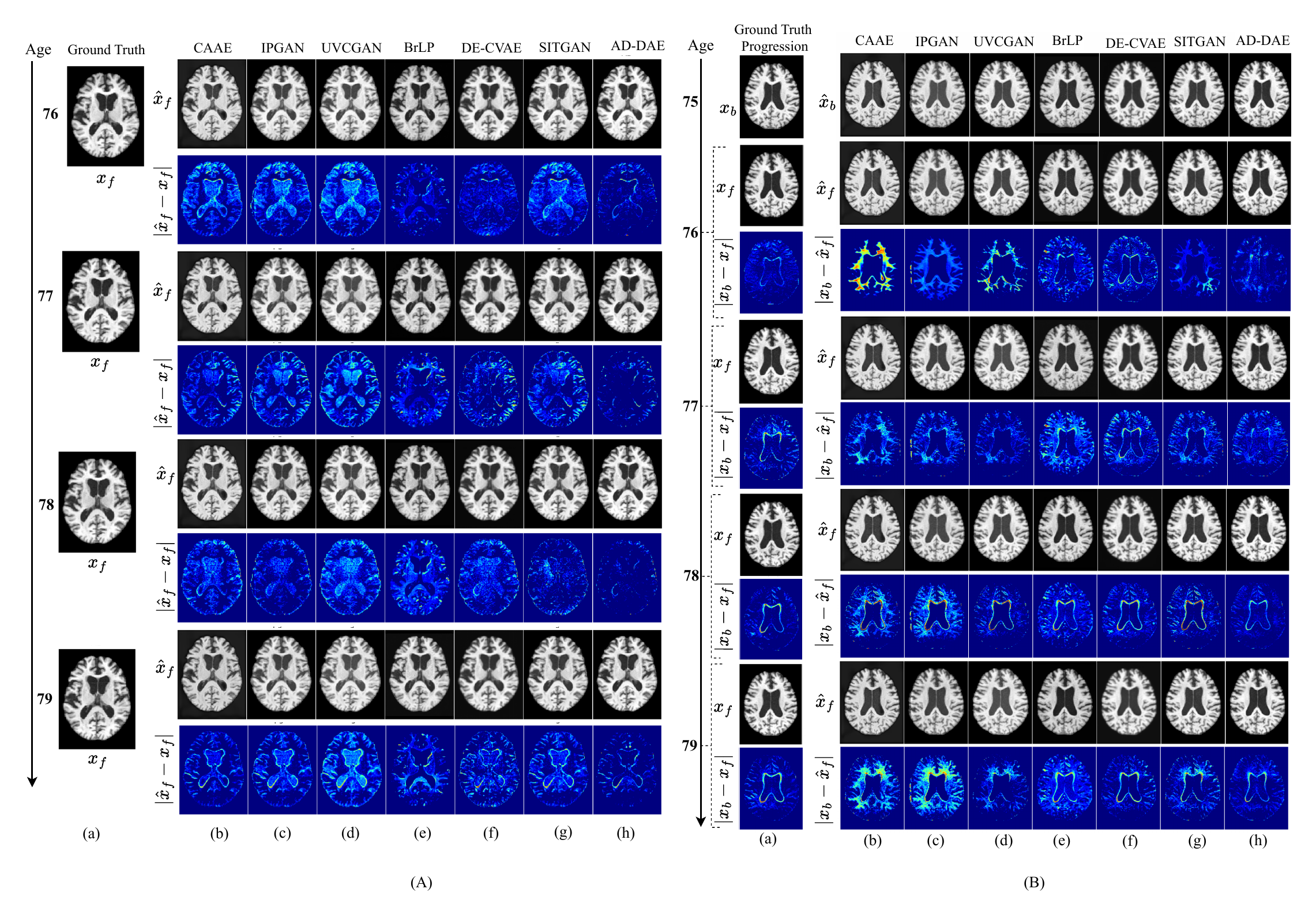}
\caption{From left to right: (A) generated follow-up images and residual maps relative to ground truth (real data from ADNI) for the MCI category (\textit{Test Set}) over a four-year age span, with column (a) showing ground-truth follow-up images and columns (b)–(h) showing predictions and residuals from AD-DAE and baseline methods (initial age 75 years). (B) previous time-point and follow-up images with residual maps illustrating four-year progression for all methods, with column (a) showing ground truth and columns (b)–(h) corresponding to AD-DAE and baselines.}
\label{fig:x_b_x_f_error_comb}
\end{figure*}

\begin{figure*}[t]
  \centering
  \includegraphics[width=1
  \linewidth]{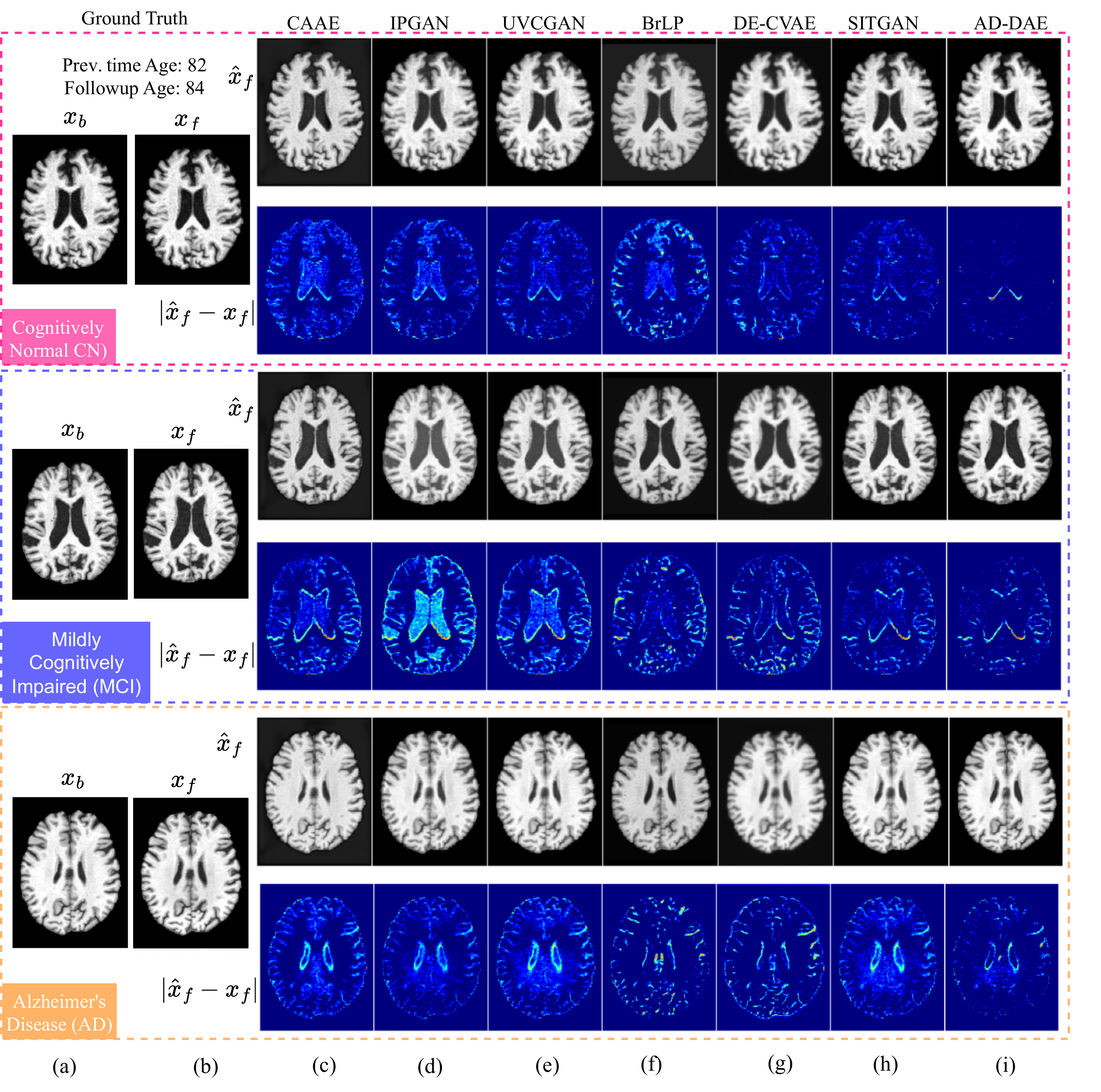}
\caption{Top to bottom: Images and error maps for CN, MCI, and AD cohorts. Left to right: (a) Ground truth previous time-point (age 82), (b) Ground truth follow-ups (age 84), (c)–(i) Generated follow-up images and error maps (relative to ground truth follow-up) for AD-DAE and baseline methods. Overall, AD-DAE shows relatively lower errors in comparison to the other methods. }
\label{fig:x_f_error_disease_wise}
\end{figure*}

\section{Results and Discussion}

The quantitative image-level results on the \textit{Test Set} and \textit{Cross-Data Setup} are presented in Subsection~\ref{subsec:Quanti}. Qualitative analyses of age- and cognitive-status–driven progression on the \textit{Test Set} are reported in Subsection~\ref{subsec:Quali}. Volumetric 3D evaluations, downstream disease classification, and latent space analyses, which validate the separation of information, are detailed in Subsections~\ref{subsec:volanalysis},~\ref{subsec:Classification}, and~\ref{subsec:latent}, respectively.

\subsection{Quantitative Analysis }
\label{subsec:Quanti}
\textit{\textbf{Performance in Test Set}}: The results of quantitative evaluation are presented in Table~\ref{table:quanti} using the image-level metrics PSNR, SSIM, and MSE. AD-DAE exhibits relatively improved performance as compared to other baselines, with PSNR gains of $1.37$ dB and $1.34$ dB and MSE reductions of $0.0002$ and $0.0003$ as compared to the better-performing baseline (SITGAN), across cognitive categories. AD-DAE also achieves a $0.01$ SSIM improvement over SITGAN for the MCI and AD categories, while showing comparable performance for CN. 
These results suggest that the image-diffusion formulation, together with explicit separation of progression-related latent components, enables better performance of AD-DAE compared to SITGAN, which does not enforce such latent separation.
Among the comparative models, the VAE–latent diffusion (BrLP) and dual-encoder VAE (DE-CVAE) approaches yield relatively lower image-level scores, as their VAE formulations introduce blurrier reconstructions that limit the ability to preserve subject-specific structures.
The GAN-based unpaired approaches (CAAE, IPGAN, UVCGAN) lack explicit separation of progression-related factors during modeling, resulting in lower overall performance. 
%
In terms of computational efficiency, AD-DAE is heavier than GAN-based approaches but remains more efficient than VAE- and diffusion-based methods.
Overall, AD-DAE \textit{\textbf{performs better}} than all the comparative baselines, demonstrating that it is more effective to model progression, ensuring that the internal latent representations \textit{\textbf{separate}} the factors of generation.

\textit{\textbf{Performance in Cross-Data Setup}}: To evaluate generalization capability, we tested the model with OASIS data in a similar setup to the \textit{Test Set} quantification and reported in Figure~\ref{fig:Oasis_plot}. Across all metrics, AD-DAE achieves relatively better or comparable performance compared to other methods.
The better-performing baseline SITGAN yields lower performance than AD-DAE, reflecting a trend similar to the ADNI \textit{Test Set} in Table~\ref{table:quanti}. In contrast to the \textit{Test Set} evaluation, BrLP performs better than DE-CVAE on all metrics in the OASIS setup, suggesting that BrLP's latent diffusion-based formulation provides better generalization than the VAE setup. For PSNR, both BrLP and DE-CVAE perform better than IPGAN and UVCGAN, whereas for MSE and SSIM, IPGAN and UVCGAN have better scores. This divergence can be attributed to the subject-specific supervision in BrLP and DE-CVAE, which, during cross-dataset evaluation, exhibit greater deviations. The performance of CAAE is relatively lower across all the scores. Overall, AD-DAE shows \textit{\textbf{better generalization}} across different cognitive states and metrics.

\subsection{Qualitative Analysis }
\label{subsec:Quali}

The qualitative comparisons of AD-DAE with other methods are shown in Figure~\ref{fig:x_b_x_f_error_comb}(A). The ground-truth and generated follow-up images, along with their differences, are presented for a four-year age gap (75–79 years). AD-DAE exhibits lower absolute error between $\hat{x}_f$ and $x_f$ compared to other approaches, reflecting improved preservation of subject-specific anatomy while capturing progression-related changes. In contrast, SITGAN shows higher errors due to limited control over progression.

Among the comparative models, DE-CVAE exhibits lower errors for smaller age gaps, but errors increase with larger progression intervals due to blurring introduced by its VAE-based formulation. BrLP similarly exhibits higher errors at larger age gaps, as its VAE-based decoder, combined with latent diffusion, provides limited control in image space, thereby reducing the preservation of subject-specific anatomy. GAN-based unpaired methods (CAAE, IPGAN, UVCGAN) capture progression less effectively due to the absence of explicit progression conditioning. Overall, AD-DAE more \textit{\textbf{effectively}} models \textit{\textbf{progression-related features}} by explicitly separating \textit{\textbf{identity}} related features in the latent space and leveraging an image-diffusion formulation.

\begin{figure*}[t]
  \centering
  \includegraphics[width=1
  \linewidth]{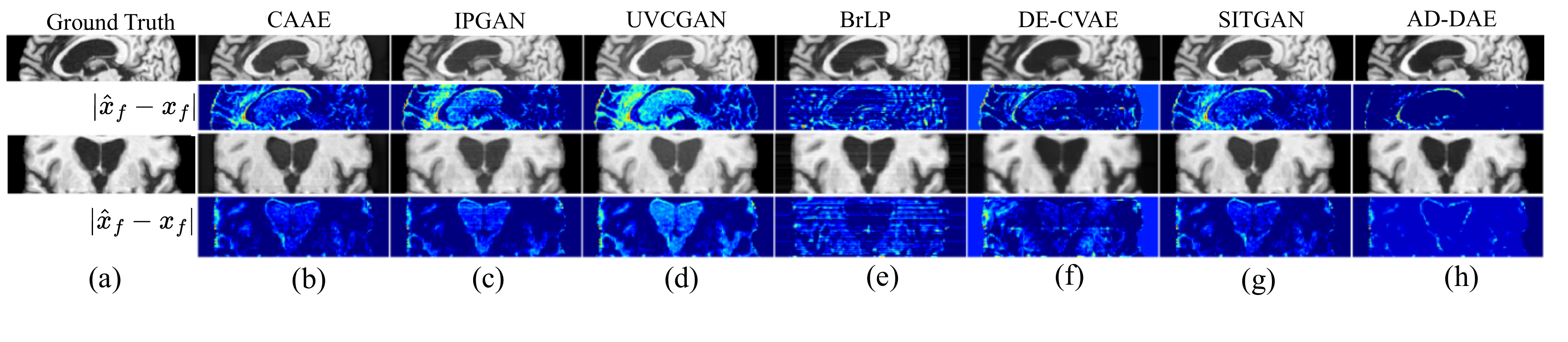}
\caption{Left to right: (a) Ground truth images of sagittal and coronal view (b)–(h) Generated sagittal and coronal view from AD-DAE and baseline methods on the \textit{Test Set} data. From top to bottom, the first and the third row represent images, while the second and the fourth highlight errors. }
\label{fig:Sag_cor_view}
\end{figure*}

\subsubsection{Anatomical Changes with Age Progression}

To assess progression modeling across increasing age, we compute difference maps between predicted/ ground truth follow-up images with the ground truth initial time-point (age 75) image for ages 76–79 across all methods (Figure~\ref{fig:x_b_x_f_error_comb}(B)). The ground-truth differences (Figure~\ref{fig:x_b_x_f_error_comb}(B)(a)) show increasing ventricular enlargement with age gap. AD-DAE closely follows this progression hierarchy, exhibiting smaller changes at lower age gaps and larger changes at higher gaps.
SITGAN captures ventricular enlargement but does not consistently preserve the progression hierarchy. DE-CVAE reflects the hierarchy but often overestimates ventricular changes due to limited control over progression factors. BrLP also follows the hierarchy; however, relatively blurrier reconstructions introduce changes beyond progression-relevant regions. In contrast, GAN-based unpaired models (CAAE, IPGAN, UVCGAN) produce diffuse, less localized changes. Overall, AD-DAE is able to \textit{\textbf{better}} capture the \textit{\textbf{progression hierarchy}}, generating \textit{\textbf{focused}} changes within \textit{\textbf{progression-relevant}} regions.

\begin{table}[]
\caption{Comparison of MAE in Anatomical Regions Across Models Changing due to Progression. Region-wise MAEs are reported on \textit{Test Set}. $^*$ indicates statistical significance ($p<0.01$).}
\centering
\resizebox{\columnwidth}{!}{%
\begin{tabular}{@{}cccc@{}}
\toprule
\textbf{Methods} & \multicolumn{3}{c}{\textbf{MAE($\downarrow$)}}                                                                          \\ \midrule
        & Hippocampus     & Amygdala        & \begin{tabular}[c]{@{}c@{}}Lateral \\ Ventricles\end{tabular} \\ \midrule
CAAE\cite{zhang2017age}    & 0.4378 $\pm$ 0.1247 & 0.3625 $\pm$ 0.0837 & 0.6535 $\pm$ 0.6484                                              \\ \midrule
IPGAN\cite{xia2021learning}   & 0.3348 $\pm$ 0.0338 & 0.3283 $\pm$ 0.0260 & 0.5445 $\pm$ 0.5240                                              \\ \midrule
UVCGAN \cite{torbunov2023uvcgan} & 0.3288 $\pm$ 0.0307 & 0.2208 $\pm$ 0.0208 & 0.5453 $\pm$ 0.3540                                               \\ \midrule
BrLP\cite{puglisi2024enhancing}    & 0.1960 $\pm$ 0.0552 & 0.1731 $\pm$ 0.0529 & 0.3702 $\pm$ 0.1012                                              \\ \midrule
DE-CVAE\cite{he2024individualized} & 0.1871 $\pm$ 0.0727 & 0.1183 $\pm$ 0.1117 & 0.1747 $\pm$ 0.1470                                              \\ \midrule
SITGAN\cite{wang2023spatial}  & 0.1161 $\pm$ 0.0288 & 0.0211 $\pm$ 0.0201 & 0.1436 $\pm$ 0.1053                                              \\ \midrule
AD-DAE  & \textbf{0.0282$^*$ $\pm$ 0.0281} & \textbf{0.0182$^*$ $\pm$ 0.0178} & \textbf{0.0405$^*$ $\pm$ 0.0391}                                              \\ \bottomrule
\end{tabular}%
}
\label{tab:vol_analysis}
\end{table}
 
\subsubsection{Progression Analysis with Cognitive States} To assess progression across cognitive statuses, we evaluate the difference in ground truth and predictions of three subjects with similar previous time-point (82 years) and follow-up (84 years) ages from different cognitive stages (Figure~\ref{fig:x_f_error_disease_wise}). AD-DAE shows consistently lower errors than other methods, with error magnitude increasing with disease severity due to larger anatomical changes. 
AD-DAE preserves subject-specific anatomy while introducing progression-correlated changes. SITGAN shows higher errors in progression-related regions despite preserving subject identity. BrLP and DE-CVAE produce spatially diffuse errors with lesser localization to progression-relevant structures, while GAN-based unpaired methods (CAAE, IPGAN, UVCGAN) do not capture the progression correlating to the cognitive categories. Overall, AD-DAE achieves \textit{\textbf{lower}} errors and models progression \textit{\textbf{consistent}} with \textit{\textbf{cognitive status variation}}.

\subsubsection{Clinical Evaluation}

The expert qualitatively interpreted that the primary atrophy patterns are captured faithfully (ventricle expansion and hippocampus shrinkage), while secondary cortical changes (grey matter thinning and sulcal widening) are observable but less pronounced. The primary indicators of disease severity remain morphologically sound.

\subsection{Analysis of Volume Synthesis}
\label{subsec:volanalysis}
To evaluate how well progression is captured in 3D, we have performed (i) Quantitative region-wise relative volume comparison (Subsection~\ref{subsubsection:Region_wise_Progression}), (ii) Qualitative analysis along the sagittal and coronal views (Subsection~\ref{subsubsection:Sagittal_Coronal_view}), and  (iii) Analysis of the deformation using Jacobian (Subsection~\ref{subsubsection:Jacobian}). Additionally, latent space analysis is more explicitly given in \href{https://github.com/ayantikadas/AD_DAE/blob/main/assets/Supplementary_Material.pdf}{Supplementary}, supporting that AD-DAE captures 2D representations that has volumetric awareness. The 2D estimated images ($\hat{x}$) are stacked, $\hat{X}_{:,:,d^{'}} = \hat{x}$ to generate the 3D images ($\hat{X} \in \mathbb{R}^{H \times W \times D}, \; d^{'} \in \{1,\dots,D\}$).

\begin{figure}[!t]
  \centering
  \includegraphics[width=1
  \linewidth]{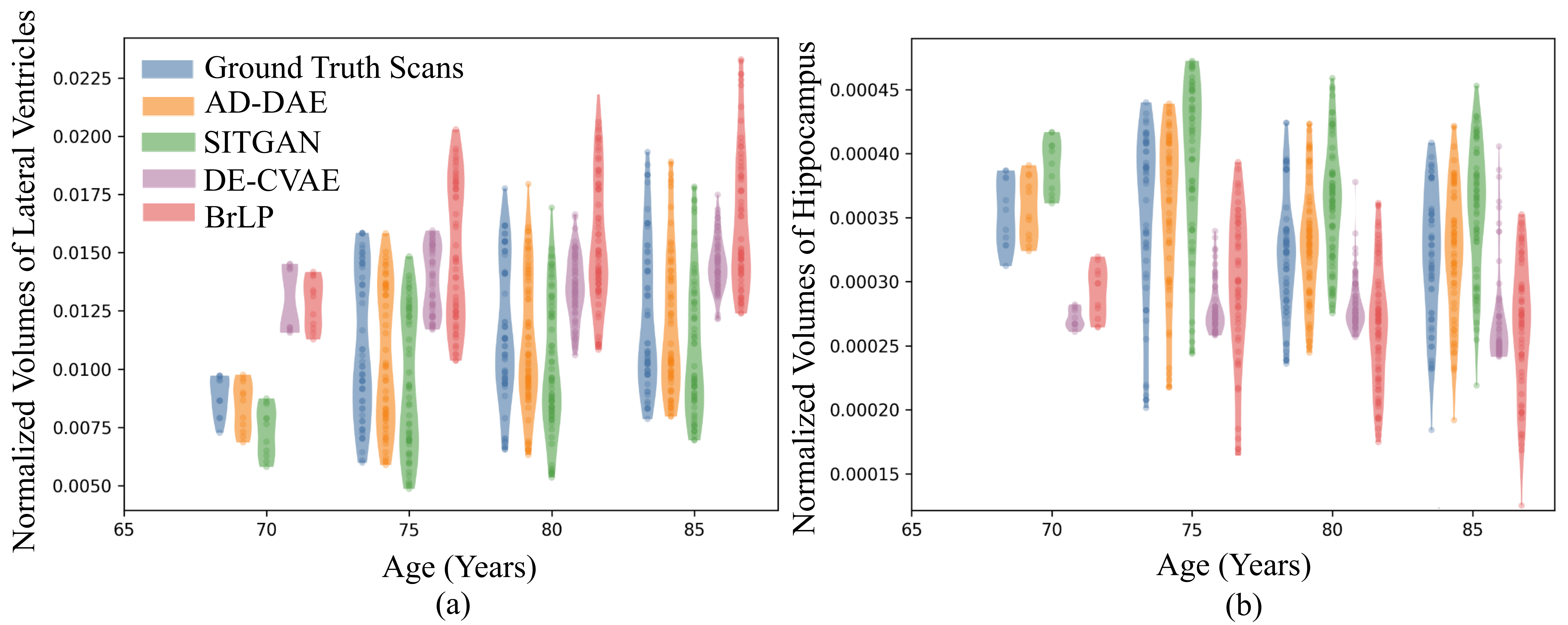}
\caption{Left to right: Normalized volumes of (a) lateral ventricles and (b) hippocampus with age for generated volumes from AD-DAE, baseline methods, and ground truth, using subjects from MCI cognitive stage (\textit{Test Set}). An overall increase in ventricular volume and a decrease in hippocampus volume are observed with age.  }
\label{fig:region_wise_progression}
\end{figure}

\subsubsection{Region-wise Progression Analysis in 3D Volumes}
\label{subsubsection:Region_wise_Progression}
The Table~\ref{tab:vol_analysis} reports region-based relative volumetric MAE between $V^{r}_{rel\_pred}$ and $V^{r}_{rel\_gt}$ using metrics as in Subsection \ref{subsec:Evaluation Metrics}.
%
From the table, AD-DAE achieves relatively lower MAE across all three regions, indicating consistent modeling of progression-related volumetric changes. The errors are generally higher in the lateral ventricles than in the hippocampus and amygdala, reflecting the greater degree of ventricular change during progression. Among baselines, SITGAN exhibits lower errors, followed by DE-CVAE and BrLP. The higher errors in DE-CVAE relative to SITGAN is due to more ventricular enlargement than the ground truth, as shown in Figure~\ref{fig:x_b_x_f_error_comb}(B)(g). BrLP shows comparatively higher error since the progression-related anatomical changes are captured in the latent space, translation to the image space does not reproduce precise image details. Overall, AD-DAE is able to better \textit{\textbf{capture}} the \textit{\textbf{progression}}-related changes at the \textit{\textbf{volumetric}} level (3D).

To further validate progression capture in 3D, normalized volumes of ground-truth ($V_{X_f}^r/V_{X_f}$) and predicted ($V_{\hat{X}_f}^r/V_{\hat{X}_f}$) follow-up ($r \in $ hippocampus and lateral ventricles) were plotted with age for MCI and AD categories using AD-DAE, and the three better-performing baseline methods (SITGAN, DE-CVAE, BrLP). The Figure~\ref{fig:region_wise_progression} (a) and (b) presents the nature of variation of the normalized volumes with ages for the hippocampus and lateral ventricles. The plots for ground-truth normalized volumes are most closely followed by AD-DAE, with SITGAN showing some deviations. Between DE-CVAE and BrLP, the former is closer to the ground truth in absolute values, whereas the latter is deviated from the ground-truth error distribution. 
For the ventricles (Figure~\ref{fig:region_wise_progression}a) and the hippocampus (Figure~\ref{fig:region_wise_progression}b), ground truth reflects enlargement and atrophy due to disease progression, a trend well followed by AD-DAE and broadly followed by other models. Overall, AD-DAE \textit{\textbf{closely follows}} the ground-truth distribution of normalized volumes.

\begin{table}[!t]
\caption{Comparison of Jacobian of Ground Truth images ($J_{x_f\rightarrow x_b}$) with the predicted images ($J_{\hat{x}_f\rightarrow x_b}$) in terms of MAE. Region-wise mean Jacobian and errors are reported on \textit{Test Set}.}
\centering
\resizebox{\columnwidth}{!}{%
\begin{tabular}{@{}lllllll@{}}
\toprule
Methods & \multicolumn{2}{c}{Hippocampus} & \multicolumn{2}{c}{Amygdala} & \multicolumn{2}{c}{Lateral Ventricle} \\ \midrule
             & $J_{Avg}$ & MAE    & $J_{Avg}$ & MAE    & $J_{Avg}$ & MAE    \\ \midrule
Ground Truth & 0.9089 & -      & 0.9642 & -      & 1.2084 & -      \\ \midrule
BrLP\cite{puglisi2024enhancing}         & 0.8778 & 0.0311 & 0.9048 & 0.0594 & 1.1125 & 0.0959 \\ \midrule
DE-CVAE\cite{he2024individualized}      & 0.9042 & 0.0047 & 0.9277 & 0.0365 & 1.1381 & 0.0703 \\ \midrule
SITGAN\cite{wang2023spatial}       & 0.9058 & 0.0031 & 0.9637 & 0.0005 & 1.1327 & 0.0757 \\ \midrule
AD-DAE       & 0.9067 & \textbf{0.0022} & 0.9636 & \textbf{0.0006} & 1.1426 & \textbf{0.0658} \\ \bottomrule
\end{tabular}%
}
\label{table:Jacobian}
\end{table}

\subsubsection{Assessing Progression Across Views (Sagittal/ Coronal)}
\label{subsubsection:Sagittal_Coronal_view}
The Figure~\ref{fig:Sag_cor_view} presents the ground truth and the generated sagittal and coronal follow-up images along with residual errors of a 77-year-old subject with MCI. The absolute error maps ($|\hat{x}_{f} - x_{f}|$) indicate that AD-DAE produces lower errors across both views. SITGAN exhibits higher errors than AD-DAE, particularly in the sagittal frontal lobe. For DE-CVAE and BrLP, errors are smaller in regions associated with progression-related changes but are higher in other structures,
These models perform better than other comparative models, as the 3D volumes are directly processed during modeling. The other comparative models (UVCGAN, IPGAN, and CAAE) show errors in both progression-related and unrelated regions. Overall, it can be inferred that AD-DAE can \textit{\textbf{better reconstruct}} the other views, showing better 2D \textit{\textbf{inter-slice consistency}}. More explicitly, \href{https://github.com/ayantikadas/AD_DAE/blob/main/assets/Supplementary_Material.pdf}{Supplementary} shows that latent representations of AD-DAE have slice-level consistent information.

\begin{figure}[]
  \centering
  \includegraphics[width=1
  \linewidth]{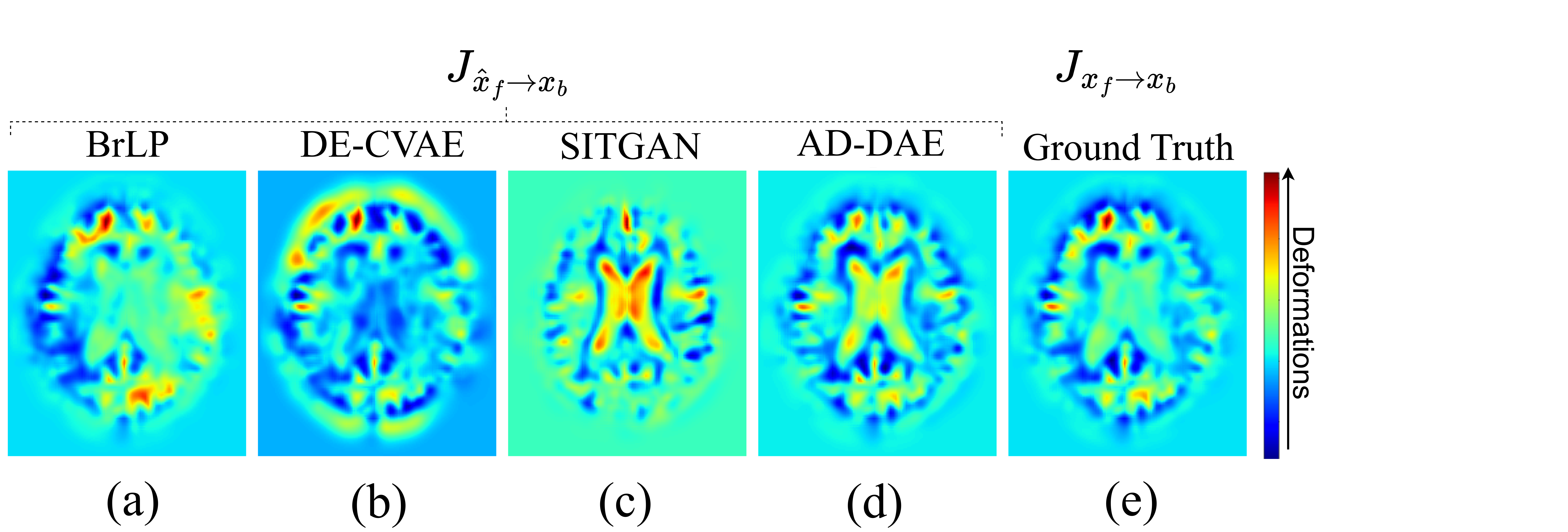}
\caption{Left to right: Jacobian of (a)-(d) predicted follow-up images ($J_{\hat{x}_f\rightarrow x_b}$) and (e) ground truth follow-up image ($J_{x_f\rightarrow x_b}$) with the previous time-point image on the \textit{Test Set} data.  }
\label{fig:Jacobian}
\end{figure}

\subsubsection{Are Anatomical Changes Captured in 3D Volumes?}
\label{subsubsection:Jacobian}
The deformations are captured by computing difference between the Jacobian determinant of generated volumes ($J_{\hat{X}_{f}\rightarrow X_{b}}$) and the ground truth ($J_{X_{f}\rightarrow X_{b}}$), as quantified in Table~\ref{table:Jacobian} in a region-wise manner. For each region, the average Jacobian determinant ($J_{Avg}$ is average of voxels-wise values in $J_{\hat{X}_{f}\rightarrow X_{b}}$ or $J_{X_{f}\rightarrow X_{b}}$) is reported. The hippocampus and amygdala exhibit $J_{Avg} < 1$, indicating atrophy, while the ventricular region shows $J_{Avg} > 1$, reflecting enlargement. All methods (BrLP, DE-CVAE, SITGAN, AD-DAE) capture atrophy and enlargement patterns, with AD-DAE more accurately reflecting ground truth progression.
Region-wise errors are highest for ventricles, followed by the hippocampus, then the amygdala, corresponding to the nature of disease progression (ventricular enlargement exceeds hippocampus atrophy, which exceeds amygdala atrophy). AD-DAE achieves relatively lower errors and $J_{Avg}$ closer to ground truth across all regions, followed by SITGAN, DE-CVAE, and BrLP. 
This highlights that the \textit{\textbf{non-linear deformations}} in the image domain are captured as \textit{\textbf{linear additive}} operation within the latent representational space of AD-DAE.

These findings are supported by Figure~\ref{fig:Jacobian}, which visualizes Jacobian determinants for ground truth and each method. AD-ADE and SITGAN exhibit more precise deformation fields, whereas DE-CVAE and BrLP produce less localized and scattered deformations. Visual representations of \textit{\textbf{deformations}} of AD-ADE are \textit{\textbf{closer}} to the ground-truth deformations.

\begin{figure}[]
  \centering
  \includegraphics[width=1
  \linewidth]{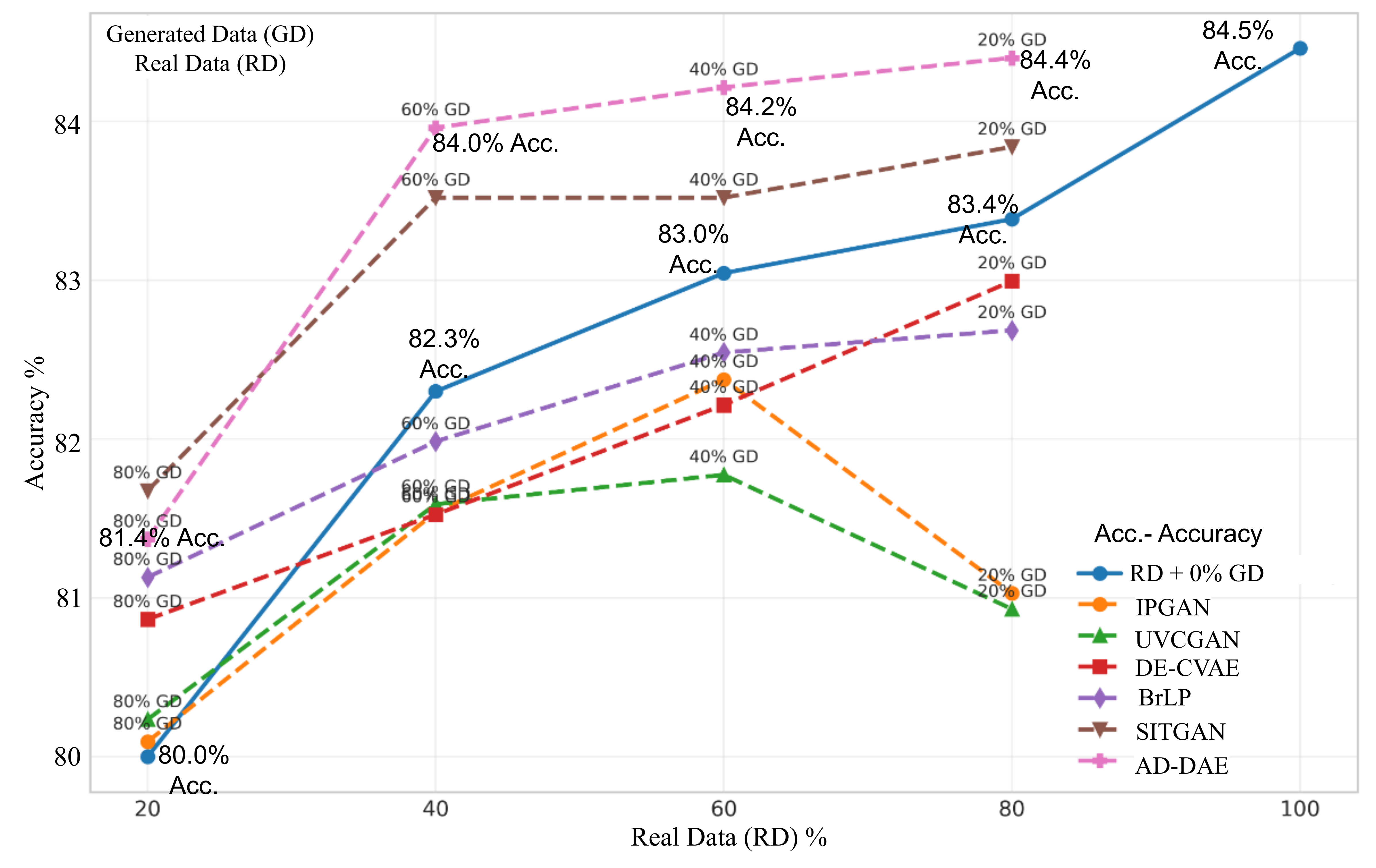}
\caption{Plots comparing disease classification accuracy of AD-DAE and the baseline methods with variable ratios of Real Data (RD) and Generated Data (GD) in the training set, with 100 RD \textit{Test Set} fixed}.
\label{fig:disease_clasification}
\end{figure}

\subsection{Downstream Analysis - Disease Classification}
\label{subsec:Classification}
To evaluate the effectiveness of generated follow-up images (GD) relative to real previous time-point ground-truth images (RD), we trained and tested Alzheimer's disease classifiers (Table~\ref{table:params}) in two different setups as described below.

\subsubsection{Generated Data Augmentation Evaluation}
In this setup, classifiers are trained with varying RD-GD ratios from the \textit{Train Set} and evaluated on RD from the \textit{Test Set} to assess how well GD augments the training process.
Figure~\ref{fig:disease_clasification} presents classification accuracy across different models from this setup. In the figure, the plots with dotted lines indicate accuracy when trained with generated data (GD) from AD-DAE, and five comparison models, and the plot with a solid line indicates training with RD-only (0\% GD). 
With RD-only (0\% GD), accuracy increases monotonically with data size. At 20\% RD $+$ 80\% GD, all models improve over the 20\% RD-only scenario, though gains vary by method. At 40\% RD $+$ 60\% GD, AD-DAE and SITGAN perform better than the RD-only (0\% GD), while other models remain below. This trend is consistent at 60\% RD $+$ 40\% GD, where AD-DAE maintains the higher accuracy compared to the other methods. 
Adding 20\% AD-DAE GD to 80\% RD improves accuracy from 83.4\% to 84.4\%, comparable to the 100\% RD accuracy of 84.5\%, indicating that AD-DAE GD have similar discriminative properties as RD.


Among the comparative models, SITGAN provides useful diversity but lower performance than AD-DAE. BrLP offers stable progression modeling but limited image-level variability, while DE-CVAE shows constrained gains due to the limited diversity of its VAE formulation. Other GAN-based models perform less favorably. Overall, AD-DAE achieves better performance across RD-GD mixtures, indicating that the \textit{\textbf{generated data effectively augments}} the training distribution.

\begin{table}[]
\centering
\caption{Classifier-based evaluation across models, trained on 100\% Real Data (RD) and tested on Generated Data (GD). The second column reports accuracy on 100\% GD from \textit{Test Set}, and the third column reports the True Positive Rate (TPR) for AD-conditioned GD from later-stage MCI.}

\label{tab:GD_test_classifier_and_MCI_AD}
\resizebox{0.75\columnwidth}{!}{%
\begin{tabular}{lcc}
\toprule
\textbf{Method} & 
\textbf{Accuracy (\%)} & 
\textbf{TPR for AD} \\
\midrule
RD Test Set          & 84.46 & 0.886 \\
\midrule
IPGAN~\cite{xia2021learning}          & 81.01 & 0.855 \\
UVCGAN~\cite{torbunov2023uvcgan}      & 81.98 & 0.856 \\
BrLP~\cite{puglisi2024enhancing}      & 83.01 & 0.865 \\
DE-CVAE~\cite{he2024individualized}   & 82.56 & 0.858 \\
SITGAN~\cite{wang2023spatial}         & 83.35 & 0.870 \\
AD-DAE                                & \textbf{84.48} & \textbf{0.879} \\
\bottomrule
\end{tabular}%
}
\end{table}

\subsubsection{Real-Trained Generated-Data Evaluation} 
\label{subsubsec:RD_train_GD_eval}
In this setup, a classifier is trained with RD from the \textit{Train Set} and evaluated on GD generated using the \textit{Test Set} to assess distributional similarity between RD and GD. The second column of Table~\ref{tab:GD_test_classifier_and_MCI_AD} reports the classification accuracy for GD from different methods. AD-DAE achieves relatively better performance than the comparative models, with accuracy closer to that obtained on the RD \textit{Test Set}. A similar trend is observed as in the first setup, where SITGAN performs relatively better than all the comparative models, and BrLP performs better than DE-CVAE due to the diversity introduced by the diffusion-based setup. Overall, these results show that AD-DAE generates GD with \textit{\textbf{higher distributional similarity}} to RD. 

\begin{figure}[!t]
  \centering
  \includegraphics[width=1
  \linewidth]{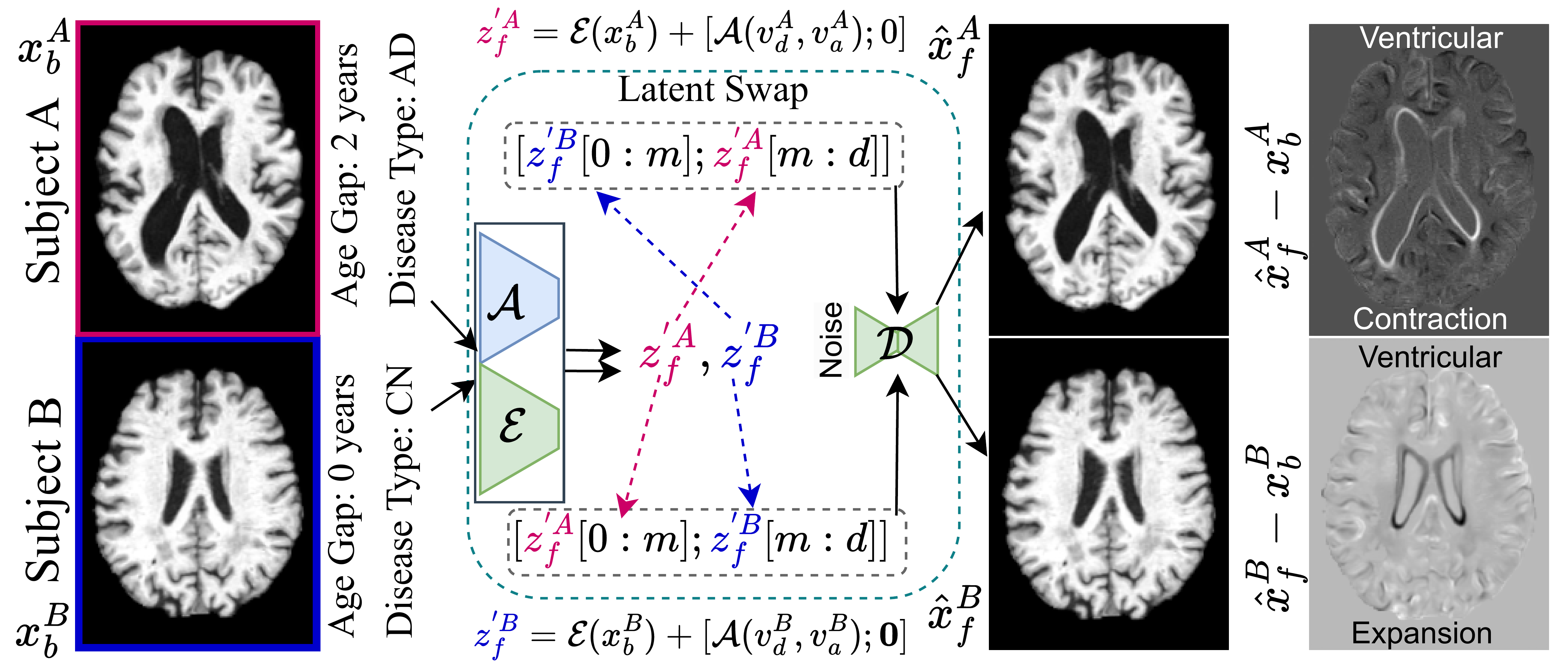}
\caption{Architectural flow for generation of images by swapping the progression-related factors of the latent representation of two subjects to assess separation of progression and subject-identity factors. Left to right: (i) Images from the two subjects, (ii) Latent generation with original cognitive status and age factors, (iii) Swapping of the 50 latent dimensions, and (iv) Generation with the swapped latents. }
\label{fig:Latent_Swap_pdf}
\end{figure}

\subsection{Cognitive State Conversion Generation}
To assess the capability of AD-DAE to generate progression across cognitive states, we generated AD-conditioned follow-up images utilizing \textit{Cross-Cognition Set} during inference. These generations were further evaluated utilizing a classifier based set-up (similar to Subsection \ref{subsubsec:RD_train_GD_eval}), since paired MCI-AD samples are limited.
%
The third column of Table~\ref{tab:GD_test_classifier_and_MCI_AD} reports the True Positive Rate (TPR) for real AD samples from the RD \textit{Test Set} and generated AD samples from different methods. 
A similar trend is observed as in Subsection~\ref{subsubsec:RD_train_GD_eval}, with AD-DAE performing relatively better, followed by SITGAN, BrLP, DE-CVAE, and the other GAN models. The TPR of AD-DAE is slightly lower than that of the RD \textit{Test Set}, since the real AD set includes both early and advanced AD cases, whereas the generated samples are synthesized from MCI inputs. Overall, AD-DAE is capable of generating MCI-to-AD progression, \textit{\textbf{capturing}} meaningful \textit{\textbf{AD-discriminative properties}}.

\begin{figure}[!t]
  \centering
  \includegraphics[width=1
  \linewidth]{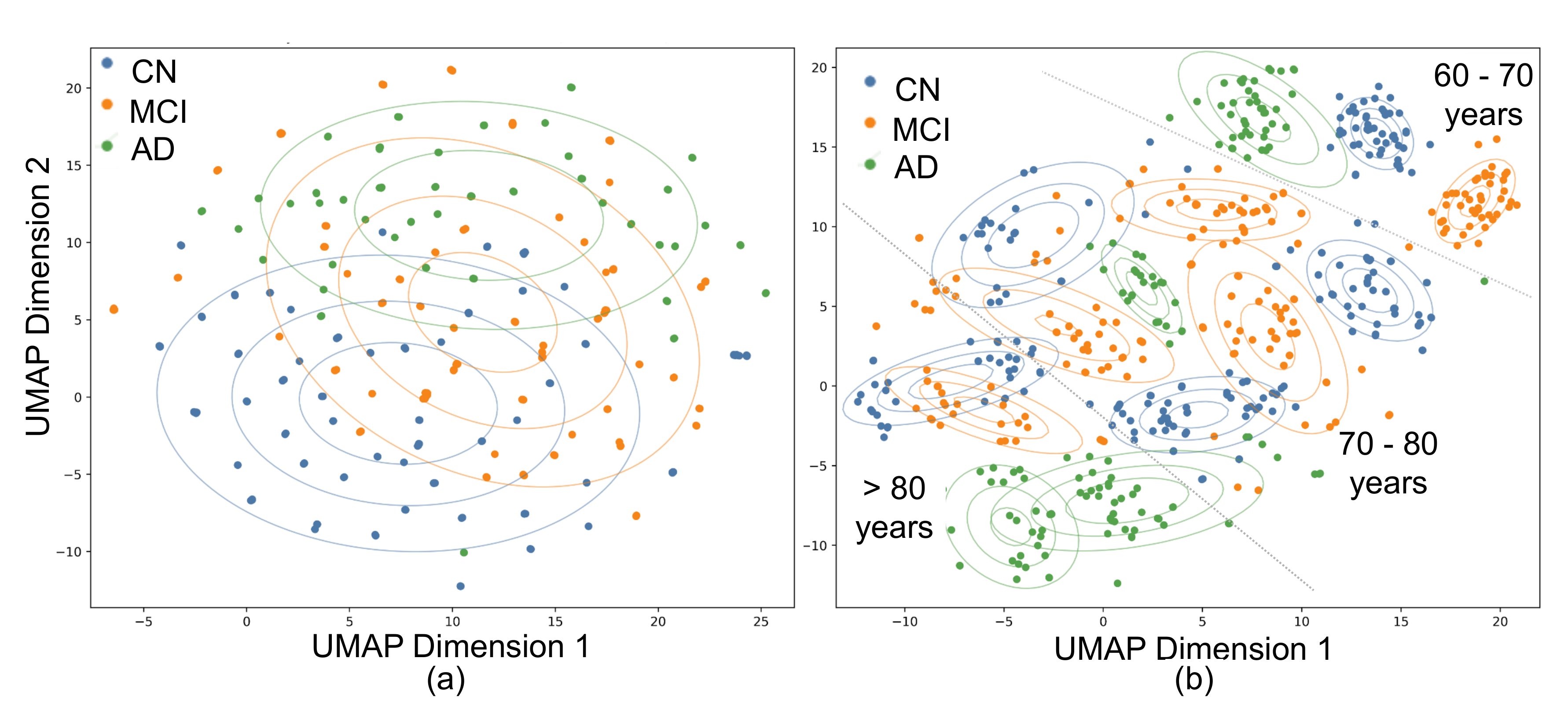}
\caption{Left to right: Umap projections of (a) the entire latent representation ($z \in \mathbb{R}^{d}$) of images, and (b) the latent factors ($[z;0],  z\in \mathbb{R}^{m}$) having the age and cognitive status-related information. }
\label{fig:Latent_Umap}
\end{figure}
\subsection{Latent Analysis }
\label{subsec:latent}
\subsubsection{Are the factors of Progression Separated?}

In order to evaluate how effectively AD-DAE separates progression-related factors from subject-specific identity, we performed a \textit{\textbf{latent swap}} experiment in which the first $m$ dimensions ($m=50$) of the latent vectors were exchanged between two subjects prior to image generation.
Utilizing the \textit{Latent Swap Dataset}, the latent vector ($z^{'B}_f =\mathcal{E}(x_b^{B}) + [\mathcal{A}(v_d^{B}, v_a^{B});0]$) of a CN subject ($x_b^{B}$) was swapped with the latent vector ($z^{'A}_f =\mathcal{E}(x_b^{A}) + [\mathcal{A}(v_d^{A}, v_a^{A});0]$) of the paired AD subject ($x_b^{A}$). The resulting follow-up for $x_b^{A}$ was generated from the modified latent $[z_{f}^{'B}[0:m]; z_{f}^{'A}[m:d]]$, yielding $\hat{x}_f^{A}$. Similarly, $[z_{f}^{'A}[0:m];z_{f}^{'B}[m:d]]$ produced a new follow-up $\hat{x}_f^{B}$. This setup for a pair of subjects is in Figure~\ref{fig:Latent_Swap_pdf}.

Qualitative results in Figure~\ref{fig:Latent_Swap_pdf} show that $\hat{x}_f^{A}$ preserves the identity of $x_b^{A}$ while exhibiting reverse progression effects, such as ventricular contraction in the residual $\hat{x}_f^{A} - x_b^{A}$. This aligns with the swapped progression attributes from $x_b^{B}$ (CN with no age gap), in contrast to the original AD subtype and two-year age gap of $x_b^{A}$. Conversely, $\hat{x}_f^{B}$ incorporates the progression attributes of $x_b^{A}$ and exhibits ventricular enlargement indicative of AD progression.
%
Overall, these results demonstrate that swapping only progression-relevant latent factors focuses on generating anatomical changes \textit{\textbf{relevant}} to \textit{\textbf{progression}}, while majorly \textit{\textbf{preserving}} subject identity. Additional quantification to validate this is given in \href{https://github.com/ayantikadas/AD_DAE/blob/main/assets/Supplementary_Material.pdf}{Supplementary}.

\subsubsection{How Latent Separation affects Organization?}


To analyze the organization of latent representations ($z$ is the average of all latents from a 3D image) learned by AD-DAE and assess the role of progression-related factors, we project $z$ into two dimensions using UMAP~\cite{mcinnes2018umap} (Figure~\ref{fig:Latent_Umap}). Projections of the complete latent vectors ($z \in \mathbb{R}^d$; Figure~\ref{fig:Latent_Umap}(a)) show overlapping clusters with limited separability, and Gaussian Mixture Model (GMM) grouping by cognitive status yields broad, intersecting ellipses. In contrast, projections of the first $m$ dimensions ($[z;0],\ z \in \mathbb{R}^m$; Figure~\ref{fig:Latent_Umap}(b)) reveal more distinct structure: clusters \textit{\textbf{align}} primarily with \textit{\textbf{cognitive statuses}} and further stratify by \textit{\textbf{age}}. GMM grouping in this subspace produces more separated ellipses, particularly for subjects aged 60-70 years, while increased overlap at higher ages reflects the combined effects of normal aging in CN and progression in AD. Overall, these results suggest that the first $m$ latent dimensions \textbf{\textit{encode age-}} and \textit{\textbf{cognition}}-related \textit{\textbf{information}} with a relevant organization.

\section{Discussion and Conclusion}

We presented \textbf{AD-DAE}, a diffusion auto-encoder framework for modeling and generating longitudinal brain MRI by encoding anatomical and progression-related information into a compact latent space and enabling condition-controlled latent shifts. The proposed formulation effectively \textit{\textbf{captures}} the brain MRI distribution and provides \textit{\textbf{precise}} control over image generation, as reflected in improved image-level metrics and volume-level analyses. Notably, volumetric results demonstrate that \textit{\textbf{non-linear}} anatomical deformations in image space can be approximated by \textit{\textbf{linear}} transformations in the latent space. Visual evaluations and latent-swap experiments further show that the explicit \textit{\textbf{separation}} of progression-related and identity-preserving factors enables meaningful anatomical changes. 

Across comparative methods, approaches that do not explicitly enforce \textit{\textbf{separation}} of progression-related factors generally benefit more from optimization using subject-specific images. While diffusion-based models better capture image distributions, latent diffusion approaches are often limited by the auto-encoding pipeline that controls reconstruction quality. Our results indicate that combining \textit{\textbf{image diffusion}} with an \textit{\textbf{auto-encoding}} formulation yields an organized latent space that supports controlled progression generation without reliance on paired longitudinal data.
\textit{Limitations and Future Scope:} The current framework conditions progression on disease status, age, and region-level information; future extensions may incorporate additional clinical factors such as beta-amyloid ($A\beta$) accumulation, treatment-related variables, and textual report–based information. Overall, diffusion autoencoders offer an effective approach for modeling longitudinal progression through condition-controlled latent transitions.

\bibliographystyle{IEEEtran}
\bibliography{AD_DAE}

\end{document}